\documentclass{article}
\usepackage{ijcai20}

\usepackage{times}
\usepackage{soul}
\usepackage{url}
\usepackage[hidelinks]{hyperref}
\usepackage[utf8]{inputenc}
\usepackage[small]{caption}
\usepackage{graphicx}
\usepackage{amsmath}
\usepackage{amsthm}
\usepackage{booktabs}
\usepackage{algorithm}
\usepackage{algorithmic}
\usepackage{makecell}
\usepackage{xspace}
\usepackage{amssymb}
\usepackage{array}
\usepackage[normalem]{ulem}
\usepackage{footmisc}
\usepackage{listings}
\usepackage{bm}
\usepackage{multirow}
\usepackage{subfigure}
\usepackage{paralist}
\usepackage{enumitem}
\usepackage{colortbl}
\usepackage{epstopdf}
\usepackage{setspace}
\usepackage{pifont}
\usepackage{xcolor}
\usepackage{mathrsfs}

\newcommand{\eg}{\emph{e.g.,}\xspace}

\newcommand{\ignore}[1]{}

\newcommand{\citet}[1]{\citeauthor{#1}~\shortcite{#1}}

\title{From Standard Summarization to New Tasks and Beyond: \\ Summarization with Manifold Information}

\author{
Shen Gao$^1$
\and
Xiuying Chen$^1$\and
Zhaochun Ren$^3$\and
Dongyan Zhao$^1$\And
Rui Yan$^{1,2}$\thanks{Corresponding Author: Rui Yan (ruiyan@pku.edu.cn)}
\affiliations
$^1$Wangxuan Institute of Computer Technology, Peking University, Beijing, China\\
$^2$Beijing Academy of Artificial Intelligence\\
$^3$School of Computer Science and Technology, Shandong University, Qingdao, China
\emails
\{shengao,xy-chen,zhaody,ruiyan\}@pku.edu.cn
zhaochun.ren@sdu.edu.cn
}

\begin{document}

\maketitle

\begin{abstract}
Text summarization is the research area aiming at creating a short and condensed version of the original document, which conveys the main idea of the document in a few words.
This research topic has started to attract the attention of a large community of researchers, and it is nowadays counted as one of the most promising research areas.
In general, text summarization algorithms aim at using a plain text document as input and then output a summary.
However, in real-world applications, most of the data is not in a plain text format.
Instead, there is much manifold information to be summarized, such as the summary for a web page based on a query in the search engine, extreme long document (\eg academic paper), dialog history and so on.
In this paper, we focus on the survey of these new summarization tasks and approaches in the real-world application.
\end{abstract}

\section{Introduction}

The rapid growth of World Wide Web means that document floods spread throughout the Internet. 
Readers get drown in the sea of documents, wondering where to access.
Text summarization system aims at generating a condensed version of a document and conveying the main idea to the reader.
Users can save a lot of time by reading the summary instead of the whole document to capture the main idea.
Hence, many websites and applications deploy automatic summarization systems, and researchers in Natural Language Processing (NLP) field also focus on the text summarization task.

Generally speaking, there are two types of text summarization. 
One is designed for the most common scenario that summarizes a plain text with hundreds of words, and the most popular usage is the news summarization.
The other one, on the contrary, is designed for generating summary with manifold information in which input may be a structured document or document with additional knowledge.
Different from the plain text summarization task, these \textit{new summarization} tasks aim to produce a better and appropriate summary by incorporating manifold information in many real-world applications and scenarios.
For example, for a search engine, it is better to summarize the web page according to the user's query instead of just summarizing the web page ignoring the query.
Another example is dialog summarization, with the development of online chatting, people always chat with people on the web for business or chitchat (\eg on Slack or Whatsapp). 
Especially in the business scenario, it is helpful to give the user a summary of what has been talked in the past days before they starting a new dialog session, or give a brief introduction to the people who just join the group chat about what has been discussed in this group.

\begin{figure}[!t]
  \centering
    \includegraphics[clip,width=0.9\columnwidth]{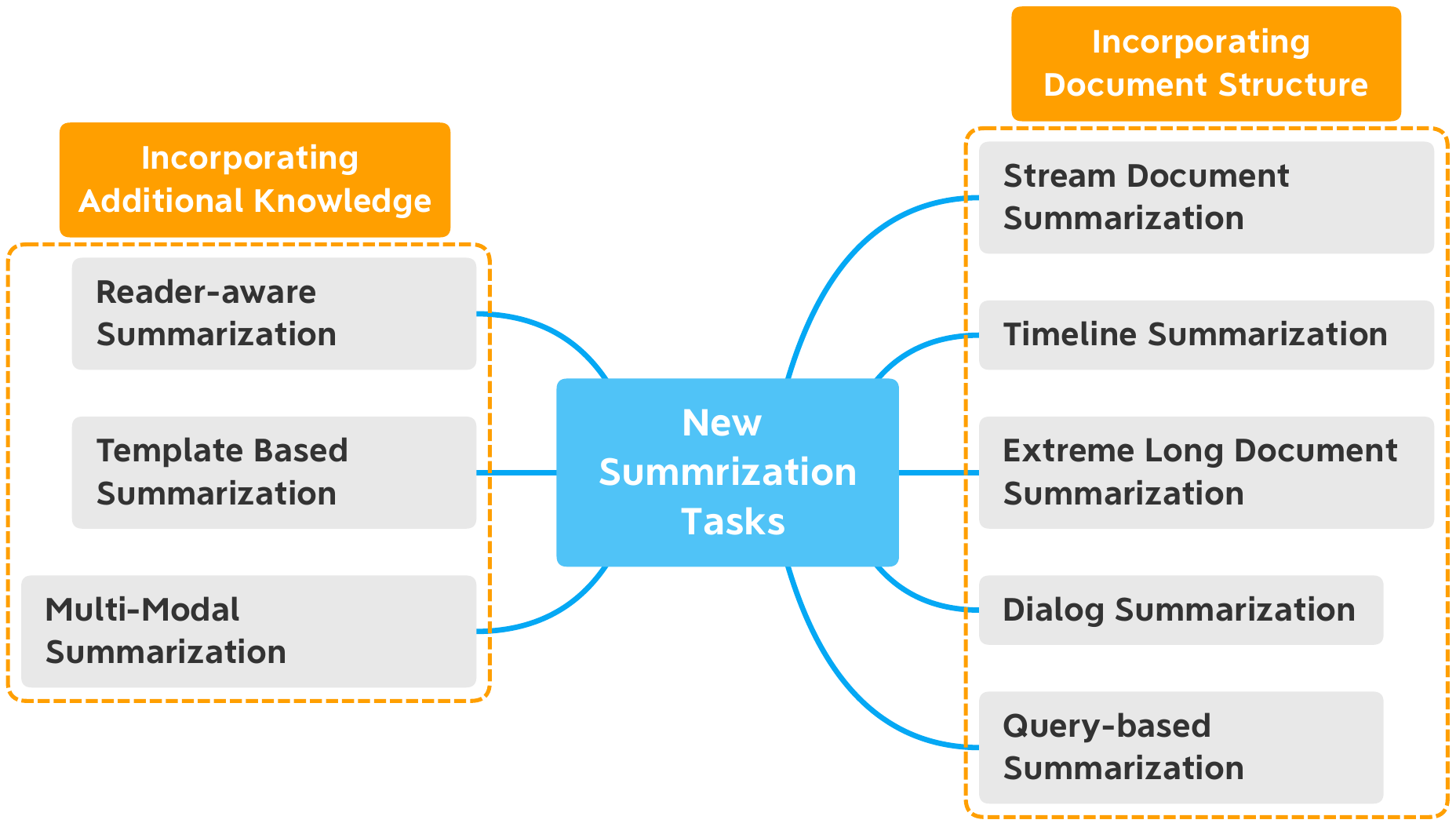}
    \caption{New summarization tasks introduced in this paper.}
  \label{fig:tasks}
\end{figure}

In contrast to the prosperity of survey on plain text summarization task, there are no systematic introductions to approaches about how to build an efficient summarization system which can leverage the manifold information (such as structure information of document and additional knowledge) in the research community. 
Thus, in this survey paper, we present a literature review for \textit{new summarization tasks} and their corresponding methods, the tasks are listed in Figure~\ref{fig:tasks}. 
There are eight summarization tasks introduced in this paper: 
(1) \textit{Stream document};
(2) \textit{Timeline document};
(3) \textit{Extreme long document};
(4) \textit{Dialog};
(5) \textit{Query-based document};
(6) \textit{Incorporating reader comment};
(7) \textit{Template based};
(8) \textit{Multi-media};
In these tasks, the first five tasks can be classified into summarization task with special \textit{incorporating document structure}. And the last three tasks can be classified into \textit{incorporating additional knowledge} into summarization.

Unlike the conventional plain text summarization methods which are built all with hand-crafted rules or feature engineering, recently many researchers begin to develop some data-driven approaches to build summarization systems.
Since these approaches can leverage the publicly available large scale dataset and the rapid progress of deep learning approaches, instead of using time-consuming hand-crafted feature engineering.
Therefore, we believe it is useful and valuable to summarize recent progress on new summarization tasks. 
This survey paper is partially based on our continuous efforts on building summarization models for new summarization tasks. 
We will introduce the problem formulation, data collection and the proposed methods for these tasks.

\begin{table}[t!]
\small
	 \centering
	 \resizebox{1.02\columnwidth}{!}{
		\begin{tabular}{lccc}
			\toprule 
			DUC & ROUGE-1 & ROUGE-2 & ROUGE-L \\ 
			\midrule
			\multicolumn{4}{@{}l}{\emph{Extractive methods}}\\
			Lead-3 & 40.42 & 17.62 & 36.67 \\ 
			\cite{Nallapati2017SummaRuNNer} & 39.60 & 16.20 & 35.30 \\
			\cite{Narayan2018Ranking} & 40.00 & 18.20 & 36.60 \\
			\cite{Wu2018Learning} & 41.25 & 18.87 & 37.75 \\
			\cite{Zhang2019HIBERTDL} & 42.37 & 19.95 & 38.83 \\
			\cite{Liu2019Text} & 43.25 & 20.24 & 39.63 \\
			\midrule
			\multicolumn{4}{@{}l}{\emph{Abstractive methods}}\\
			\cite{See2017Get} & 39.53 & 17.28 & 36.38 \\ 
			\cite{Paulus2018A} & 39.87 & 15.82 & 36.90 \\
			\cite{Hsu2018A} & 40.68 & 17.97 & 37.13 \\
			\cite{Celikyilmaz2018Deep} & 41.69 & 19.47 & 37.92 \\
			\cite{Liu2019Text} & 42.13 & 19.60 & 39.18 \\
			\bottomrule
		\end{tabular}
		}
	\caption{\label{tab:performance} Leaderboard of document summarization task on CNN/DailyMail dataset.}
\end{table}

\section{Preliminary: Standard Summarization}

In this section, we will introduce some generally used summarization frameworks based on conventional and neural-based learning methods.
These frameworks are used as the basis of the methods for new summarization tasks.

\subsection{Conventional Methods}

Early conventional approaches to extractive summarization include: 
Centroid-based methods~\cite{Radev2004CentroidbasedSO,Lin2002FromST}, 
supervised and semi-supervised methods~\cite{Wong2008ExtractiveSU}, 
tree and graph based methods~\cite{Kikuchi2014SingleDS,Qian2013FastJC,Morita2013SubtreeES,Filippova2008DependencyTB},  
Submodular methods~\cite{Morita2013SubtreeES,Li2012MultidocumentSV,Lin2010MultidocumentSV} and
ILP-based methods~\cite{Gillick2009ASG,Li2013UsingSB,Banerjee2015MultiDocumentAS}.
Nevetheless, extractive approaches only extract some phrases or sentences from the original document as the summary, and it can not produce the condensed and fluent summary~\cite{Yates2007TextRunnerOI}.
On the contrary, the conventional abstractive summarization methods usually extract some words from document, and then reorder and perform linguistically-motivated transformations to the words~\cite{Dorr2003HedgeTA,Banko2000HeadlineGB,Cohn2009SentenceCA,Barzilay2005SentenceFF,Tanaka2009SyntaxDrivenSR}.
However, these paraphrase-based generation method are easy to produce influent sentences.

\subsection{Neural Methods}

\begin{figure}[!t]
  \centering
    \includegraphics[clip,width=0.9\columnwidth]{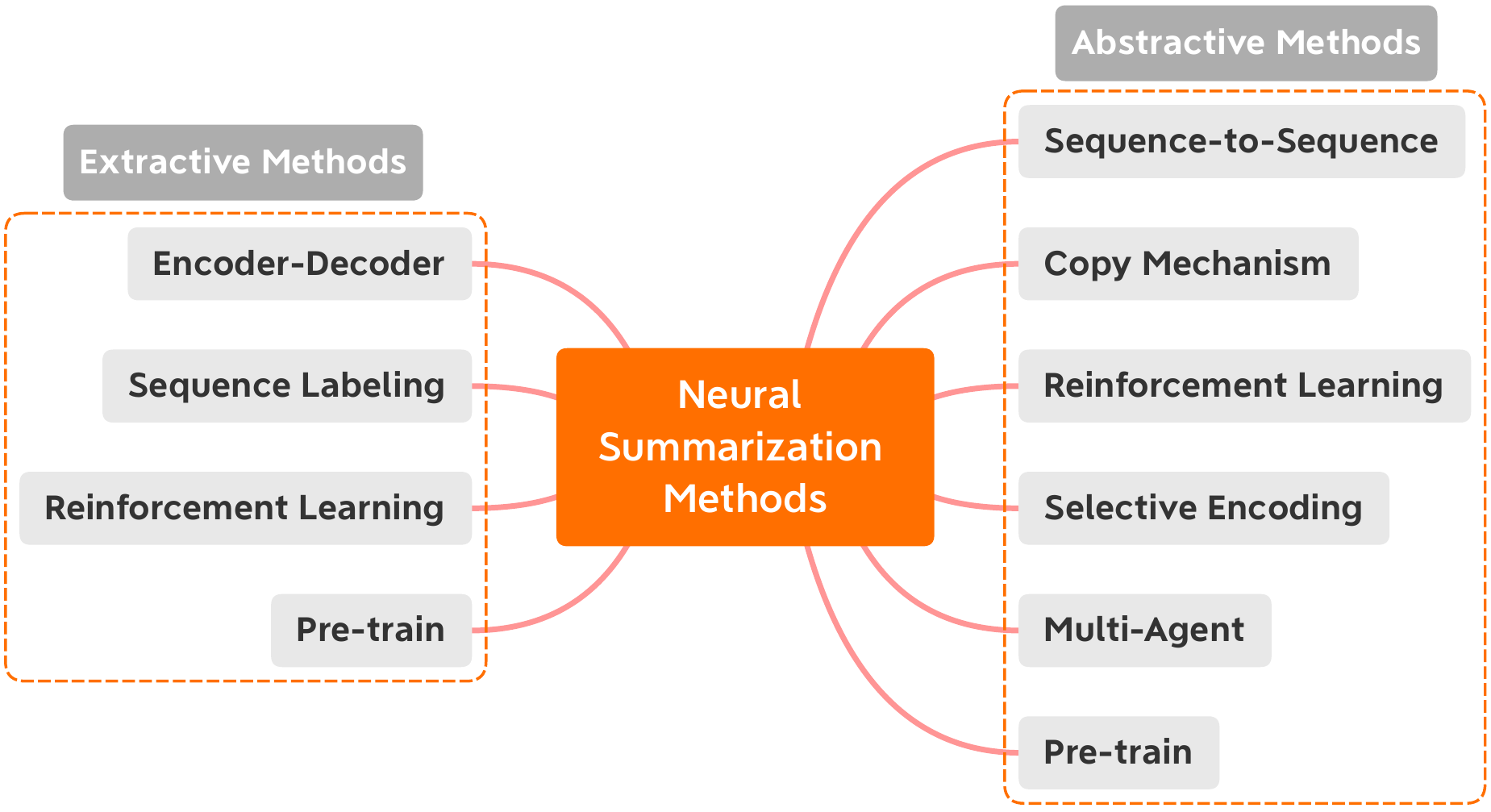}
    \caption{Techniques used in neural-based summarization methods.}
  \label{fig:neural-summ}
\end{figure}

In contrast to the conventional learning methods, neural-based approaches provide an end-to-end method to summarization task.
In this section, we introduce some widely used techniques in these methods, as shown in Figure~\ref{fig:neural-summ}, and we split the these methods into two categories: extractive and abstractive.
Most of these works conduct the experiments on a benchmark dataset CNN/DailyMail, and we list the performance in terms of ROUGE score~\cite{lin2004rouge} at Table~\ref{tab:performance}.

First, we will introduce the extractive methods which extract sentences from the document as the summary.
Since the extractive methods use the sentence as the basic unit, the first step is to obtain a sentence representation.
The most common way~\cite{cheng2016neural,Narayan2018Ranking} is to employ a recurrent neural network (RNN) or convolutional neural network (CNN) to encode the words in a sentence, and then obtain a vector representation.
After obtaining the sentence representation, \citet{cheng2016neural} first propose a framework with \textit{encoder-decoder} using an RNN to tackle the extractive summarization task, which uses the encoder to obtain the vector representation of sentences and use the decoder with an attention mechanism to extract sentence.
Since the \textit{encoder-decoder} summarization framework~\cite{Narayan2018Ranking} needs two RNN that computes slowly, many researchers start to use the \textit{sequence labeling} framework~\cite{Nallapati2017SummaRuNNer,Zhou2018NeuralDS} for this task which use an RNN to read the sentences only once.
To deeply understand the document, researchers incorporate the memory network~\cite{Chen2018IterativeDR} into summarization framework which gives the reasoning ability to the model.
Since previous works use the cross-entropy as the loss function to train the model which has a gap with the testing stage that use the ROUGE~\cite{lin2004rouge} score, \citet{Wu2018Learning} propose to use the \textit{reinforcement learning} method to directly optimize the ROUGE score. 
In recent two years, the \textit{pre-training} techniques growth rapidly in NLP field.
Researchers employ the language model pre-training model (like Elmo, BERT) into summarization task~\cite{Liu2019Text,Zhang2019HIBERTDL} to achieve better performance.
From Table~\ref{tab:performance}, we can find that \cite{Liu2019Text} achieve the state-of-the-art performance on the CNN/DailyMail dataset.


The \textit{sequence-to-sequence} based text generation methods~\cite{Sutskever2014SequenceTS,Li2019Learning,Gao2019ProductAware} make generating fluent and concise summary possible, and \citet{Nallapati2016AbstractiveTS} firstly apply the text generation method to the abstractive summarization task.
Next, many extensions on the text generation framework are proposed to achieve better performance in generating summary.
To avoid the out-of-vocabulary (OOV) problem, \textit{copy mechanism}~\cite{Gu2016,See2017Get,Gulcehre2015} is proposed which directly copy the OOV words (like the name of person, place or institution) from the source document into the generated summary. 
Similarly, researchers also use the \textit{reinforcement learning} method in abstractive summarization~\cite{Paulus2018A,Wang2018A,Liu2018Generative} for the same reason as the extractive methods.
To help the summary generation module focus on the salient parts of source document, \textit{selective encoding}~\cite{Zhou2017Selective,Hsu2018A} are proposed to encode the important semantic parts and ignore the trivial parts.
Encoding the source document is a crucial step in summarization, researchers~\cite{Celikyilmaz2018Deep} propose to divide the hard document encoding task into several sub-tasks and solve it by \textit{multiple collaborating encoder agents}.
After encoding the document, an attention mechanism is used to fuse all the document representations produced by all the agents, and then generate the summary.
\textit{Pre-training} language model helps the model to capture the semantic of text, that motivates the researchers to use the it as the encoder of summarization framework.
Researchers employ the pre-training technique into document reading module~\cite{Liu2019Text} to capture the main idea of the document, and achieve the state-of-the-art performance in abstractive summarization task.
These methods use a plain text as input, and they can not utilize the document structure or other easily acquired knowledge to improve the summarization performance.

\section{Summarization by Incorporating Document Structure}

In this section, we will introduce some summarization tasks in which input is structural text instead of a plain text.
These special structures will help the summarization model to capture the document main idea.

\subsection{Stream Document Summarization}

Stream summarization task was first introduced in TAC 2008~\cite{Dang2008OverviewOT}, which targets at summarizing new documents in a continuously growing \textit{text stream} such as news events and twitters.
When the new document arrives in a sequence, the stream summary needs to be updated along with, considering previous information meantime.

Initial works include~\cite{Boudin2008ASM}, which proposes a scalable sentence scoring method for query-oriented update summarization.
In this method, candidate sentences are selected according to a combined criterion of query relevance and dissimilarity with previously read sentences.
\citet{Delort2012DualSumAT} present an unsupervised probabilistic approach to model novelty in a document collection and apply it to the generation of update summaries.
\citet{Ge2016NewsSS} propose a graph ranking based method, Burst Information Networks, as a novel representation of a text stream.
In this method, the graph node is a burst word (including entities) with the time span of one of its burst periods, and an edge between two nodes indicates how strongly they are related.
\cite{Mnasri2017TakingIA} is the state-of-the-art work on the DUC and TAC dataset, which examines how integrating a semantic sentence similarity into an update summarization system can improve its results.

Current state-of-the-art methods for this task are all based on human-engineered and extractive methods~\cite{Hong2014ARO,Mnasri2017TakingIA}.
Nowadays, there are many stream data provided on the internet, such as tweets focus on the same news topic and news of a big event (like a presidential election, a natural disaster).
Consequently, the abstractive-based summarization methods will be a hot research area and will be explored in the near future.

\subsection{Timeline Summarization}

Classic news summarization plays an important role with the exponential document growth on the Web. 
Many approaches are proposed to generate summaries but seldom simultaneously consider evolutionary characteristics of news plus to traditional summary elements.
Timeline summarization is an important research task which can help users to have a quick understanding of the overall evolution of any given topic. 
It thus attracts much attention from research communities in recent years. 
To solve this task, one should first identify which sub-events are salient and then generate a summary.
The big difference between timeline summarization and stream summarization task is whether the model can see all the sub-event.

Timeline summarization task is firstly proposed by~\citet{allan2001temporal}, in this paper, they propose a method that extracts a single sentence from each event within a news topic.
Later, a series of works~\cite{yan2011evolutionary,yan2011timeline,yan2012visualizing,zhao2013timeline} further investigate the timeline summarization task, and all of them are based on conventional learning method to extract sentences from the timeline data.
For instance, \cite{yan2011evolutionary} formulate the timeline summarization task as a balanced optimization problem via iterative substitution.
The objective function used in this method is measured by four properties: relevance, coverage, coherence, and diversity.
In recent years, as an important case of timeline information, social media data is used by many timeline summarization research works.
For example, \citet{ren2013personalized} considered the task of time-aware tweets summarization, based on a user’s history and collaborative social influences from ``social circles''.

The previous works are all based on extractive methods, which are not as flexible as abstractive approaches.
\citet{Chen2019Learning} firstly propose a key-value memory network-based architecture to store the events described in the timeline.
In this key-value memory network, they use the event time representation as the key, and split the value into two slots: global value and local value.
The local value only captures event information from current event and the global value stores the global characteristics of events in different time position.
Finally, an RNN-based decoder is employed to generate the summary abstractively.
To train their model, they release the first large-scale timeline summarization dataset which contains 179,423 document-summary pairs collected from a cyclopedia website.

\subsection{Extreme Long Document Summarization}

Different from the previous summarization task, in some scenarios, the input document can be very long, such as an academic paper or a patent document which is longer than the news article.
Thus, summarizing such an extreme long document is still a challenging problem when using existing summarization methods.
The biggest challenge of this task is to extract the salient information and central idea from a large amount of information.

First, we introduce some benchmark datasets used the extreme long document summarization.
In the era of using conventional machine learning methods, in this task, all of the researchers use the small-scale scientific papers as the dataset~\cite{Teufel2002SummarizingSA,Teufel2002SummarizingSA,Liakata2013ADC}, and the number document summary data pairs is less than 100.
In recent years, most of the researchers employ the neural-based methods which require a large amount of data to train the model.
Thus, many large-scale long document datasets have been proposed and the data comes from Wikipedia~\cite{Liu2018Generating}, scientific papers~\cite{Cohan2018ADA}, patent documents~\cite{Sharma2019BIGPATENT}, etc.
\citet{Liu2018Generating} propose to use a Wikipedia web page with all the reference articles and the results fetched from Google as the long text input, and there are 2,332,000 document summary data pairs in this dataset.
\cite{Cohan2018ADA} propose a large-scale scientific paper summarization dataset which is collected from arXiv and PubMed, and it contains 348,000 document and summary pairs.
The average document length is 4938 and 3016 words in arXiv and PubMed respectively, which is $6$ times longer than the news dataset CNN/DailyMail.
\cite{Sharma2019BIGPATENT} propose a larger long document summarization dataset BigPatent, which is $10$ times larger than the PubMed and arXiv.
It contains 1,341,362 US patent document and summarization pairs and the average document length is 3,572 words.
This dataset is more suitable for abstractive summarization task since the summary has more novel n-grams than other long document summarization datasets.

In the initial works of this task, researchers~\cite{Teufel2002SummarizingSA,Liakata2013ADC} use a supervised classifier to select content from a scientific paper based on some human-engineered features. 
Then, researchers have extended these works by applying more sophisticated classifiers to identify more fine-grain categories.
To determine whether a sentence should be included in the summary, \cite{collins2017supervised} directly use the section each sentence appears in as a categorical feature with values like Highlight, Abstract, Introduction, etc.

As for the neural network based methods, \citet{Liu2018Generating} firstly use an extractive summarization method to coarsely identify salient information and then employ a neural abstractive model to generate the summary.
\cite{Cohan2018ADA} propose a hierarchical model that uses two-level RNN to encode the words and sections respectively, and then they use an attention decoder to forms a context vector from both word and sentence level information.
They also conduct experiments on arXiv and PubMed datasets, and their model outperforms the baseline methods on these datasets.
\cite{Xiao2019Extractive} propose an extractive method for this task using both the global context of the whole document and the local context within the current topic, and this method achieves state-of-the-art performance on the previous two datasets.

\subsection{Dialog Summarization}

In recent years, online chatting becomes more and more popular~\cite{Qiu2019Are,Tao2018Learning,Gao2020Learning}.
When the chatting history becomes very long, it is time-consuming for people to review all the context before starting a new dialog.
Thus, some researchers focus on the task of summarizing the dialog history.
Different from summarizing a document, the salient information is scattered in the whole dialog history.

\cite{Ganesh2019AbstractiveSO} first propose this task and they propose a pipeline method that consists of a sequence labeling module to identify the salient utterance and a Seq2Seq module with attention and copy mechanism. 
Since their dataset is in a small-scale, they use a news summarization dataset CNN/DailyMail to train the abstractive summarization module and evaluate on a small scale dialog summarization dataset with only 45 sessions.
To leverage the neural-based text generation method, \cite{Gliwa2019SAMSum} propose the first large scale dataset SAMSum for this task.
Different from previous papers working on chit-chats, \cite{Liu2019Automatic} propose a framework to generate a summary for online customer service, which can help the staff to know what was happen without going through long and sometimes twisted utterances.

As another branch of dialog summarization task, the meeting summarization task is to generate a summary of meeting transcriptions.
\cite{Shang2018Unsupervised} propose an unsupervised abstractive meeting summarization using a graph-based model and budgeted submodular maximization.
In recent years, people usually hold a meeting using video calls instead of just using audio.
Consequently, additional visual information can be used in meeting summarization, such as the participant's head pose and eye gaze.
\cite{Li2019Keep} propose a multi-modal encoding framework that incorporates this visual information and employs a topic segmentation method to identify the topic transition in a dialog flow.
Finally, they employ the Pointer-Generator~\cite{See2017Get} network to fusion the encoded information and generate the summary.

\subsection{Query-based Summarization}

In the typical web search scenario, the search engine provides a list of web pages associated with their summaries.
Different from the traditional document summarization, in this scenario, the summary should summarize the query focused aspect of the web page instead of the main idea.
Inspire by this application, many researchers start to focus on the query-based summarization task, whose goal is to generate a summary that highlights those points that are relevant in the context of a given query.

In this task, most of the methods are based on conventional machine learning methods.
\cite{li2014query} propose a semi-supervised graph-based model and incorporate the LDA topic model into summarization.
\cite{Feigenblat2017Unsupervised} propose an unsupervised multi-document query-based summarization method using a cross-entropy method which is a generic Monte-Carlo framework for solving hard combinatorial optimization problems.
Different from previous sentence extraction methods, \cite{wang2013sentence} employ a sentence compression method which uses three compression strategy: rule-based, sequence-based and tree-based to produce the summary.

To avoid generating repeated phrases and increasing the diversity of summary, \cite{Nema2017Diversity} firstly propose a neural-based Seq2Seq framework which ensures that context vectors in attention mechanism are orthogonal to each other.
Specifically, to alleviate the problem of repeating phrases in the summary, we treat successive context vectors as a sequence and use a modified LSTM cell to compute the new state at each decoding time step.
In decoding steps, the attention mechanism is also used to focus on different portions of the query at different time steps.

\section{Summarization by Incorporating Additional Knowledge}

In some summarization applications, there are many different types of additional knowledge that can be used to help the model enhance the performance.
The model can leverage this additional knowledge to capture the main idea of the document or generate more fluent summaries.

\subsection{Reader-aware Summarization}

In most of the news websites, they provide an area for the readers to post their comments on the news article.
In most cases, the reader comments concentrate on the main idea of the news article.
Thus, these comments can be used to help the summarization model to capture the main idea of the news, and then the model can generate a better summary with this help.
In this section, we will introduce two kinds of methods which are based on conventional learning methods and neural networks respectively.

In the beginning, \cite{Hu2008CommentsorientedDS} firstly propose to understand the input document with the feedback of readers using a graph-based method, where they identify three relations (topic, quotation, and mention) by which comments can be linked to one another.
\cite{Li2015ReaderAwareMS} employ a sparse coding based framework for this task which jointly considers news documents and reader comments via an unsupervised data reconstruction strategy.

Next, we turn to the methods using neural networks.
\cite{Li2017ReaderAwareMS} propose a sentence salience estimation framework based on a neural generative model called Variational Auto-Encoder (VAE). 
In contrast to the previous methods which use sentence extraction methods on a small-scale dataset, \cite{Gao2019Abstractive} first propose a large-scale dataset and a neural generative method RASG on this task.
This dataset contains 863,826 data samples, and each data sample has several a document, a summary and several reader comments (the average comments number of a document is 9.11).
The proposed RASG method is a generative-adversarial~\cite{goodfellow2014generative} based learning method which conducts the interaction between the reader comments and news article to capture the reader attention distribution on the article, and then use the reader focused article information to guide the summary generation process. 

\subsection{Template Based Summarization}

To generate a fluent summary,
template based summarization method first retrieves a summary template and then edits it into the new summary of the current document.
Existing methods can be classified into two categories: hard-editing and soft-editing.
More specifically, hard-editing methods force the system to generate the summary which is in the same language pattern as the template.
Conversely, soft-editing methods can use partial words in the template and generate more flexible summaries.

\cite{wang2013domain} introduced a template-based focused abstractive meeting summarization system. 
Their system first applies a MultipleSequence Alignment algorithm to generate templates and extracts all summary-worthy relation instances from the document. 
Then, the templates are filled with these relation instances. 
\cite{Oya2014ATA} propose a hard-editing method that employs a multi-sentence fusion algorithm in order to generate summary templates.

Since the hard-editing methods are not very flexible, soft-editing methods become popular in recent years due to the development of the neural text generation method.
\cite{Cao2018Retrieve} employ existing summaries as soft templates to generate a new summary.
In this method, they use an information retrieval system to retrieve summaries of a similar document and then use an attention-based generator to fuse the information from the template and current document.
\cite{wang2019biset} leverages template discovered from training data to softly select key information from each source article to guide its summarization process.
However, this method ignores the dependency between the template document and the input document.
Following these works, \cite{Gao2019How} propose to analyze the dependency and use this dependency to help the model identify which facts in the input document are the salient facts that should be mentioned in the summary.
Furthermore, they use the relationship between template document and template summary to extract the summary template that can be reused in generating a new summary.
In \cite{Gao2019How}, they also propose a large-scale dataset (contains 2,003,390 document and summary pairs) in which summaries are all in patternized language, and their method achieves state-of-the-art performance on this dataset.

\subsection{Multi-Modal Summarization}

With the increase of multi-media data on the web, many researchers focus on the multi-modal summarization task~\cite{Zhu2018MSMOMS,Li2018Multimodal,Chen2018Abstractive,Li2017Multimodal,Palaskar2019Multimodal,Li2019Read} in recent years.
Compared to the traditional text summarization task setting, in the multi-modal summarization task, the visual information is incorporated along with the input document into the text summarizing process to improve the quality of the generated summary.

In the beginning, we first introduce some datasets of image-based multi-modal summarization.
\cite{Li2018Multimodal,Chen2018Abstractive} propose two large-scale multi-modal summarization datasets, and each data sample in these datasets contains a source sentence, an image collected from the webpage and a summary.
Different from the previous two datasets which output is only text, \cite{Zhu2018MSMOMS} propose the first large-scale multi-modal input and multi-modal output summarization dataset which input is a document with several images and the ground truth is a text summary with the most relevant image selected from the input images.
\cite{Li2017Multimodal,Palaskar2019Multimodal} propose two video-based multi-modal summarization dataset which contains document, video and summary pairs, and the number of data sample are 1000 and 79114 respectively.

Next, we will introduce some existing methods of multi-modal summarization task.
For the image-based multi-modal summarization task, \cite{Li2018Multimodal,Chen2018Abstractive,Zhu2018MSMOMS} propose to use a Seq2Seq based abstractive model which has image and document encoders to obtain the representations of multi-modal input and an RNN based decoder with multi-modality attention to generate the summary. 
Since there are some abstract concepts in the source document which can not find a counterpart in the image, and not all the visual information is useful for generating summary.
To avoid introducing noise into summarization, \cite{Li2018Multimodal} propose to use an image attention filter and an image context filter.
As for the video-based summarization, \cite{Palaskar2019Multimodal} employ a ResNeXt-101 3D~\cite{Hara2017CanS3} convolutional neural network to model the video frames and then fuse this video information into the Seq2Seq using a hierarchical attention mechanism.

\section{Conclusion}

We have witnessed a rapid surge of summarization studies recently, especially the research in the many new summarization tasks. 
Summarization systems are catching on fire: the state-of-the-art performance of summarization tasks has been pushed higher and higher.
In the real-world applications, most of them are not in a traditional summarization setting, and they usually leverage manifold information. 
Since large-scale big data become more easily available in our living era, and it requires much time for people to obtain the overall information for an event.
We may stand at the entrance of future success in more advanced summarization systems.
It is our hope that this survey provides an overview of the challenges and the recent progress as well as some future directions in these new summarization tasks which leverages the manifold information.


\section*{Acknowledgements}

We would like to thank the anonymous reviewers for their constructive comments. 
This work was supported by the National Key Research and Development Program of China (No. 2017YFC0804001), the National Science Foundation of China (NSFC No. 61876196 and NSFC No. 61672058).
Rui Yan is partially supported as a Young Fellow of Beijing Institute of Artificial Intelligence (BAAI).

\bibliographystyle{named}
\bibliography{new-summ-survey}

\end{document}